\documentclass[10pt,leqno]{amsart}

\usepackage{graphicx}
\usepackage{indentfirst,csquotes}
\usepackage{natbib}
\usepackage{amssymb,amsthm,amsmath}
\usepackage{xcolor,paralist,hyperref,titlesec,fancyhdr,etoolbox}
\usepackage{lipsum}
\usepackage{listings}
\lstnewenvironment{example}{}{}
\makeatletter
\providecommand{\@chapapp}{\relax}
\makeatother
\makeatletter
\providecommand\@secnumpunct{\quad} 
\makeatother

\newtheorem{thm}{Theorem}[section]

\newcommand{\argmin}{\operatornamewithlimits{argmin}}
\newcommand{\mb}{\mathbb}      
\newcommand{\vc}{\mathbf}      

\newcommand{\mc}{\mathcal}
\newcommand{\mr}{\mathrm}

\titleformat{\section}[display]
  {\normalfont\huge\bfseries\centering}
  {\centering}{10pt}{\Large}
\titlespacing*{\section}{0pt}{0ex}{0ex}

\hypersetup{colorlinks=true, linkcolor=black, filecolor=black, urlcolor=black}

\begin{document}
\title{RKUM:   An R    Package for Robust Kernel Unsupervised  Methods} 
\author[Initial Surname]{Md Ashad Alam}
\date{\today}
\address{Centers for Outcomes and Health Services Research, Ochsner Health, 1514 Jefferson Highway, New Orleans, LA, USA}
\email{example@mail.com}
\maketitle

\begin{abstract}
RKUM is an R package developed for implementing robust kernel-based unsupervised methods. It provides functions for estimating the robust kernel covariance operator (CO) and the robust kernel cross-covariance operator (CCO) using generalized loss functions instead of the conventional quadratic loss. These operators form the foundation of robust kernel learning and enable reliable analysis under contaminated or noisy data conditions. The package includes implementations of robust kernel canonical correlation analysis (Kernel CCA), as well as the influence function (IF) for both standard and multiple kernel CCA frameworks. The influence function quantifies sensitivity and helps detect influential or outlying observations across two-view and multi-view datasets. Experiments using synthesized two-view and multi-view data demonstrate that the IF of the standard kernel CCA effectively identifies outliers, while the robust kernel methods implemented in RKUM exhibit reduced sensitivity to contamination. Overall, RKUM provides an efficient and extensible platform for robust kernel-based analysis in high-dimensional data applications. 
\end{abstract} 

\section{Introduction}
\label{sec:Intro}
Although substantial progress has been made in developing robust methods for supervised learning, such as support vector machines for classification and regression \cite{Christmann-04,Christmann-07,Debruyne-08}, comparatively few principled approaches exist for kernel-based unsupervised learning. Robustness remains a critical yet challenging aspect of applying statistical machine learning to multi-source and high-dimensional data, where the presence of outliers can severely distort analysis and inference. Since the 1960s, researchers in robust statistics have proposed various approaches aimed at reducing the influence of outlying observations. The central goal of these methods is to extract meaningful structure from the majority of the data while identifying deviations from the expected pattern \cite{Huber-09,Hampel-11, Ashad-08}.

More recently, robust kernel density estimation (robust kernel DE) methods have been introduced, leveraging the concept of robust kernel mean elements (robust kernel ME) to reduce sensitivity to outliers \cite{Kim-12}. These estimators are optimized within a reproducing kernel Hilbert space (RKHS) using a kernelized iteratively reweighted least squares (KIRWLS) algorithm. In parallel, two spatially robust kernel principal component analysis (PCA) approaches have been developed—one based on weighted eigenvalue decomposition \cite{Huang-KPCA, Ashad-13,Ashad-11, Ashad-14T} and another employing spherical kernel PCA \cite{Debruyne-10, Alam-16a, Alam-16b,  Alam-18C}. Both studies demonstrated that the influence function (IF) of conventional kernel PCA, a key indicator of robustness, can become unbounded when non-compact kernels are used, underscoring the need for robust kernel formulations.

Most kernel-based methods explicitly or implicitly rely on the kernel covariance operator (CO) and the kernel cross-covariance operator (CCO), which serve as fundamental components in unsupervised kernel learning. However, these operators have traditionally lacked robust formulations. In this work, we demonstrate that both operators can be expressed as empirical optimization problems, enabling robust estimation by integrating the principles of Huber and Hampel within the M-estimation framework \cite{Huber-09,Hampel-11, Alam-19}. The proposed robust kernel CO and robust kernel CCO can be efficiently estimated using a kernelized iteratively reweighted least squares (KIRWLS) algorithm, ensuring stability in the presence of outliers and contaminated data.

In this work, we present RKUM, an R package developed to implement robust kernel-based unsupervised learning methods. The package provides efficient tools for estimating the robust kernel covariance operator (CO) and robust kernel cross-covariance operator (CCO) using generalized loss functions, extending beyond the conventional quadratic loss to enhance robustness against outliers and data contamination. It also includes implementations of robust kernel canonical correlation analysis (kernel CCA) and the influence function (IF) for both standard and multi-kernel CCA frameworks. Through synthetic two-view and multi-view data experiments, we demonstrate that the proposed robust estimators significantly reduce sensitivity to outliers while preserving the ability to detect influential observations. Overall, RKUM provides a unified and extensible platform for conducting robust, high-dimensional, and multi-source data analyses using kernel-based unsupervised learning techniques.

\section{Background and methods}

Let $F_X$, $F_Y$  and $F_{XY}$ be probability measures on the given nonempty sets $\mc{X}$, $\mc{Y}$ and  $ \mc{X}\times\mc{Y}$, respectively,  such that $F_X$ and $F_Y$ are the marginals of $F_{XY}$. Also let $X_1, X_2, \ldots, X_n$;  $ Y_1, Y_2, \ldots, Y_n$ and $(X_1, Y_1),(X_2, Y_2), \ldots, (X_n, Y_n)$ be the independent and identically distributed (IID) samples from the distribution  $F_X$, $F_Y$  and $F_{XY}$, respectively. A symmetric kernel, $k(\cdot,\cdot)\colon \mc{X}\times \mc{X} \rightarrow \mb{R}$, defined on a space  is called a {\bf positive definite kernel} if the Gram matrix $(k(X_i, X_j))_{ij}$ is positive semi-definite for all $i,j \in \{1, 2, \cdots, n\}$. A RKHS is a Hilbert space with a reproducing kernel whose span is dense in the Hilbert space. We can equivalently define an RKHS as a Hilbert space of functions with all evaluation functionals bounded and linear. The Moore-Aronszajn theorem states that every symmetric, positive definite kernel defines a unique reproducing kernel Hilbert space \cite{Aron-RKHS}. The {\bf feature map} is a  mapping $ \vc{\Phi}: x \to \mc{H}_X$ and  defined as $\vc{\Phi}(x) = k(\cdot, x), \forall\, x\in \mc{X}$). The vector $\vc{\Phi}(x) \in \mc{H}_X$ is called  a {\bf feature vector}. The inner product of two feature vectors can be defined as  $\langle  \vc{\Phi} (x), \vc{\Phi} (x^\prime)\rangle_{\mc{H}_X} = k(x, x^\prime)$ for all $x, x^\prime \in \mc{X}$. This is called  the {\bf kernel trick}. By the reproducing property, $f(x)=\langle f(\cdot), k(\cdot, x)\rangle_{\mc{H}_X}$, with   $f \in \mc{H}_X$  and the kernel trick, the kernel can evaluate the inner product of any two feature vectors efficiently, without knowing an explicit form of either the feature map or the feature vector. Another great advantage is that the computational cost does not depend on the dimension of the original space after computing the Gram matrices \cite{Fukumizu-14,Ashad-14T, Richfield-17,Ashad-13,Ashad-16,ALAM2021,afroz2024multi,Zhu-24,Imtyaz2025}.

\subsection{Robust kernel mean element}
\label{sec:me}
Let $k_X$ be a measurable positive definite kernel on $\mc{X}$ with $\mb{E}_X[\sqrt{k(X, X)}] < \infty$.  The {\bf kernel mean}, $\mc{M}_X$, of $X$ on  $\mc{H}_X$ is an element of $\mc{H}_X$ and is defined by the mean of the $\mc{H}_X$-valued random  variable $k_X(\cdot, X)$,
\[ \mc{M}_X(\cdot)=\mb{E}_X[k_X(\cdot, X)].\]
The kernel mean always exists with arbitrary probability  under the assumption that positive definite kernels are bounded and measurable.
By the reproducing property, the kernel ME satisfies the following equality
 \[\langle \mc{M}_X, f \rangle_{\mc{H}_X} = \langle \mb{E}_X[k_X(\cdot, X)], f \rangle_{\mc{H}_X} = \mb{E}_X\langle  k_X(\cdot, X), f \rangle_{\mc{H}_X}=\mb{E}_X[f(X)],\] for all    $f\in \mc{H}_X$.

 The  empirical kernel ME, $\widehat{\mc{M}}_X=\frac{1}{n}\sum_{i=1}^n\vc{\Phi}(X_i)=  \frac{1}{n} \sum_{i=1}^n k_X(\cdot, X_i)$ is an  element of the RKHS,
\[\langle \widehat{\mc{M}}_X, f \rangle_{\mc{H}_X}= \langle  \frac{1}{n}\sum_{i=1}^n k_X(\cdot, X_i), f\rangle =  \frac{1}{n}\sum_{i=1}^n  f(X_i).\]
The  empirical kernel ME of  the feature vectors $\vc{\Phi}(X_i)$  can be regarded as a solution to the empirical risk optimization problem \citep{Kim-12}
\begin{eqnarray}
\label{EROP1}
\widehat{\mc{M}}_X=\argmin_{f\in \mc{H}_X} \sum_{i=1}^n\| \vc{\Phi}(X_i)- f\|^2_{\mc{H}_X}.
\end{eqnarray}

The   kernel ME is the  solution to the empirical risk optimization problem, which is a least square type of  estimator.  This type of estimator is sensitive  to the presence of outliers in the feature, $\vc{\Phi}(X)$.  To reduce the effect of outliers, we can use $M$-estimation. In recent years, the robust kernel ME has been proposed for density estimation \citep{Kim-12}. The robust kernel ME,  based on  a robust loss function  $\zeta(t)$ on $t \geq 0$, is defined as
\begin{eqnarray}
\label{EKME1}
\widehat{\mc{M}}_R=\argmin_{f\in \mc{H}_X} \sum_{i=1}^n\zeta(\| \vc{\Phi}(X_i)- f\|_{\mc{H}_X}).
\end{eqnarray}
Examples of robust loss functions include Huber's loss function,  Hampel's loss function, or Tukey's biweight loss function. Unlike the quadratic loss function, the derivative of these loss functions are bounded \citep{Huber-09, Hampel-86, Tukey-77,Hassan-21,Auliah-21}. The  Huber's function, a hybrid approach between squared and absolute  error losses, is defined as:
\begin{eqnarray}
\zeta(t)=
\begin{cases}
	 t^2/2,\qquad  \qquad 0\leq t\leq c
\\
 ct-c^2/2,\qquad c\leq t, \nonumber
\end{cases}
\end{eqnarray}
where c ($c >0$) is a tuning parameter. The Hampel's loss function is defined as:
\begin{eqnarray}
\zeta(t)=
\begin{cases}
	 t^2/2,\qquad  \qquad\qquad\qquad 0\leq t\le c_1
\\
 c_1t-c_1^2/2,\qquad\qquad  c_1\leq t < c_2
\\
-\frac{c_1}{2(c_3-c_2)}(t-c_3)^2+ \frac{c_1(c_2+c_3-c_1)}{2},\qquad  c_2\leq t < c_3\\
 \frac{c_1(c_2+c_3-c_1)}{2}, \qquad\qquad  c_3\leq t, \nonumber
\end{cases}
\end{eqnarray}
where the non-negative free parameters $c_1 < c_2 < c_3$ allow us to control the degree of suppression of large errors. The Tukey's biweight loss functions is defined as:
\begin{eqnarray}
\zeta(t)=
\begin{cases}
	 1-(1-(t/c)^2)^3,\qquad  \qquad 0\leq t\leq c
\\
 1,\qquad c\leq t, \nonumber
\end{cases}
\end{eqnarray}
where $c >0$.

The basic assumptions of the loss functions are; (i) $\zeta$ is non-decreasing, $\zeta(0)=0$ and  $\zeta(t)/t \to 0$ as $t\to 0$, (ii) $\varphi(t)=\frac{\zeta^\prime(t)}{t}$  exists and is finite, where $\zeta^\prime(t)$ is the derivative of $\zeta(t)$, (iii) $\zeta^\prime(t)$ and $\varphi(t)$ are continuous, and bounded, and  (iv) $\varphi(t)$ is Lipschitz  continuous. All of these assumptions hold for Huber's loss function as well as others \citep{Kim-12}.

Essentially  Eq. (\ref{EKME1}) does not have a closed form solution, but using KIRWLS, the solution of robust kernel mean is,
\[\widehat{\mc{M}}_R^{(h)}= \sum_{i=1}^n w_i^{(h-1)}k_X(\cdot, X_i),\]
where $w_i^{(h)}=\frac{\varphi(\|\tilde{\vc{\Phi}}(X_i) - f^{(h)}\|_{\mc{H}_X})}{\sum_{b=1}^n\varphi(\| \vc{\Phi}(X_b)- f^{(h)}\|_{\mc{H}_X})}\,, \rm{and} \, \varphi(x)=\frac{\zeta^\prime(x)}{x}.$

Given the weights of the robust kernel ME, $\vc{w}=[ w_1, w_2, \cdots, w_n]^T$, of  a set of observations $X_i, \cdots, X_n$,  the points $\tilde{\vc{\Phi}}(X_i):= \vc{\Phi}(X_i) - \sum_{a=1}^nw_a \vc{\Phi}(X_a)$  are centered and the centered robust Gram matrix is $\tilde{K}_{ij}= \langle \tilde{\vc{\Phi}}(X_i),\tilde{\vc{\Phi}}(X_j)\rangle= (\vc{C}\vc{K}_X\vc{C}^T)_{ij}$,
where  $\vc{K}_X = (k_X (X_i, X_j))_{i=1}^n$ is a Gram matrix, $\vc{1}_n=[1_1, 1_2, \cdots, 1_n]^T$ and $\vc{C}=\vc{I}- \vc{1}_n\vc{w}^T$.
For a set of test points $ X^t_1, X_2^t, \cdots, X_T^t$,  we define two matrices of order $T\times n$ as  $K_{ij}^{test}= \langle \vc{\Phi}(X^t_i), \vc{\Phi}(X_j) \rangle$ and  $\tilde{K}_{ij}^{test}= \langle \vc{\Phi}(X^t_i)- \sum_{b=1}^nw_b \vc{\Phi}(X_b), \vc{\Phi}(X_j)- \sum_{d=1}^nw_d  \vc{\Phi}(X_b)\rangle$.  Like the centered Gram matrix,  the centered robust Gram matrix of test points,  $K_{ij}^{test}$,  in terms of  the robust Gram matrix and $\vc{1}_t=[1_1, 1_2, \cdots, 1_t]^T$  is defined as,
\begin{eqnarray}
\label{CM2}
\tilde{K}_{ij}^{test} =  (\vc{K}^{test}- \vc{1}_t\vc{w}^T \vc{K} - \vc{K}^{test} \vc{w}\vc{1}_n^T + \vc{1}_t\vc{w}^T \vc{K} \vc{w}\vc{1}_n^T)_{ij} \nonumber
\end{eqnarray}

\subsection{Standard kernel cross-covariance operator}
\label{sec:kernel-CCO}

In this section, we study the covariance between two random feature mappings $k_X(\cdot, X)$ and $k_Y(\cdot, Y)$. 
Analogous to the covariance of standard random vectors, the kernel cross-covariance operator (CCO) provides a framework to quantify statistical dependence between two variables in reproducing kernel Hilbert spaces (RKHSs).

Let $(\mc{X}, \mc{B}_X)$ and $(\mc{Y}, \mc{B}_Y)$ be measurable spaces, and let $(X, Y)$ be a random variable on $\mc{X} \times \mc{Y}$ with joint distribution $F_{XY}$. 
The centered kernel CCO is a linear operator $\Sigma_{XY}: \mc{H}_Y \to \mc{H}_X$ defined by
\[
\Sigma_{XY} = \mb{E}_{XY}\!\left[\tilde{\vc{\Phi}}(X) \otimes \tilde{\vc{\Phi}}(Y)\right],
\]
where $\tilde{\vc{\Phi}}(\cdot)=\vc{\Phi}(\cdot)- \mb{E}[\vc{\Phi}(\cdot)]$, and $\otimes$ denotes the tensor product operator. 
For any $a_1, b_1 \in \mc{H}_1$ and $a_2, b_2 \in \mc{H}_2$, the tensor product satisfies 
\[
(a_1 \otimes b_1)x = \langle x, b_1 \rangle a_1, \quad
\langle a_1 \otimes a_2, b_1 \otimes b_2 \rangle_{\mc{H}_1 \otimes \mc{H}_2}
= \langle a_1, b_1 \rangle_{\mc{H}_1} \langle a_2, b_2 \rangle_{\mc{H}_2},
\]
where $\mc{H}_1$ and $\mc{H}_2$ are Hilbert spaces \citep{Reed-80}.

Assume $k_X$ and $k_Y$ are measurable, positive definite kernels with corresponding RKHSs $\mc{H}_X$ and $\mc{H}_Y$, and that $\mb{E}_X[k_X(X, X)] < \infty$ and $\mb{E}_Y[k_Y(Y, Y)] < \infty$. 
Then, by the reproducing property, for any $f_X \in \mc{H}_X$ and $f_Y \in \mc{H}_Y$,
\begin{align*}
\langle f_X, \Sigma_{XY} f_Y \rangle_{\mc{H}_X}
&= \mb{E}_{XY}\!\left[
\langle f_X, k_X(\cdot, X) - \mc{M}_X \rangle_{\mc{H}_X}
\langle f_Y, k_Y(\cdot, Y) - \mc{M}_Y \rangle_{\mc{H}_Y}
\right]\\
&= \mb{E}_{XY}\!\left[
(f_X(X) - \mb{E}_X[f_X(X)]) (f_Y(Y) - \mb{E}_Y[f_Y(Y)])
\right].
\end{align*}
Thus, $\Sigma_{XY}$ is a bounded linear operator.

Similar to Eq.~\eqref{EROP1}, the kernel CCO can be expressed as an empirical risk minimization problem:
\begin{equation}
\label{EROP2}
\widehat{\Sigma}_{XY}
= \argmin_{\Sigma \in \mc{H}_X \otimes \mc{H}_Y}
\sum_{i=1}^n 
\big\| \tilde{\vc{\Phi}}(X_i) \otimes \tilde{\vc{\Phi}}(Y_i) - \Sigma \big\|^2_{\mc{H}_X \otimes \mc{H}_Y}.
\end{equation}

The empirical kernel CCO can be written as
\begin{align*}
\widehat{\Sigma}_{XY} 
&= \frac{1}{n} \sum_{i=1}^n 
\left(k_X(\cdot, X_i) - \frac{1}{n} \sum_{b=1}^n k_X(\cdot, X_b)\right)
\otimes
\left(k_Y(\cdot, Y_i) - \frac{1}{n} \sum_{d=1}^n k_Y(\cdot, Y_d)\right)\\
&= \frac{1}{n} \sum_{i=1}^n 
\tilde{k}_X(\cdot, X_i) \otimes \tilde{k}_Y(\cdot, Y_i),
\end{align*}
where $\tilde{k}_X$ and $\tilde{k}_Y$ denote the centered kernels. 
In the special case where $Y = X$, $\widehat{\Sigma}_{XY}$ reduces to the empirical kernel covariance operator.

\section{Robust kernel (cross-) covariance operator}
\label{sec:RKCCO}
Because a  robust kernel ME (see Section \ref{sec:rkme}) is used, to reduce the effect of outliers, we propose to use $M$-estimation  to find a robust sample covariance  of  $\vc{\Phi}(X)$ and $\vc{\Phi}(Y)$.  To do this, we  estimate kernel CO and kernel CCO  based on robust loss functions, namely,  robust kernel CO and  robust kernel CCO, respectively.  Eq. (\ref{EROP2}) can be written as
\begin{eqnarray}
\label{REROP1}
\widehat{\Sigma}_{RXY}= \argmin_{\Sigma\in {\mc{H}_X\otimes\mc{H}_Y}} \sum_{i=1}^n \zeta(\| \tilde{\vc{\Phi}}(X_i) \otimes \tilde{\vc{\Phi}}(Y_i) - \Sigma\|_ {\mc{H}_X\otimes\mc{H}_Y}).
\end{eqnarray}

\section{Robust kernel canonical correlation analysis}
Canonical correlation analysis  is a method for identifying the effective-dependent variables and measure the relationships between two multivariate variables, all measured on the same individual \cite{Hotelling-36}. For example, it can be used to identify how genetic markers such as SNPs are associated with fMRI voxels of a patient.

In this section, we  review standard kernel CCA and propose the IF  and empirical IF (EIF) of kernel CCA. After that we propose a {\bf robust kernel CCA} method based on robust kernel CO and robust kernel CCO.

\subsection{Standard kernel canonical correlation analysis}
\label{sec:CKCCA}
 Standard kernel CCA has been proposed as a nonlinear extension of linear CCA \citep{Akaho}. Researchers have extended the standard kernel CCA  with  an efficient computational algorithm, i.e., incomplete Cholesky factorization \cite{Back-02}.  Over the last decade,  standard kernel CCA has been used for various tasks \citep{Alzate2008,Huang-2009, Ashad-15}. Theoretical results on the convergence of kernel CCA have also been  obtained \citep{Fukumizu-SCKCCA,Hardoon2009}.

The aim of the  standard kernel CCA is to seek  the sets of functions in the RKHS for which the correlation (Corr) of  random variables is maximized. For the simplest case, given two sets of random variables $X$  and $Y$ with  two  functions in the RKHS, $f_{X}(\cdot)\in \mc{H}_X$  and  $f_{Y}(\cdot)\in \mc{H}_Y$, the optimization problem of  the random variables $f_X(X)$ and $f_Y(Y)$ is
\begin{eqnarray}
\label{ckcca1}
\rho =\max_{\substack{f_{X}\in \mc{H}_X,f_{Y}\in \mc{H}_Y \\ f_{X}\ne 0,\,f_{Y}\ne 0}}\mr{Corr}(f_X(X),f_Y(Y)).
\end{eqnarray}
The optimizing functions $f_{X}(\cdot)$ and $f_{Y}(\cdotp)$ are determined up to scale.

Using a  finite sample, we are able to estimate the desired functions. Given an i.i.d sample, $(X_i,Y_i)_{i=1}^n$ from a joint distribution $F_{XY}$, by taking the inner products with elements or ``parameters" in the RKHS, we have features
$f_X(\cdot)= \sum_{i=1}^na_{Xi}k_X(\cdot,X_i) $ and
 $f_Y(\cdot)=\sum_{j=1}^na_{Yj}k_Y(\cdot,Y_j)$, where $k_X(\cdot, X)$ and $k_Y(\cdot, Y)$ are the associated kernel functions for $\mc{H}_X$ and $\mc{H}_Y$, respectively. The kernel Gram matrices are defined as   $\vc{K}_X:=(k_X(X_i,X_j))_{i,j=1}^n $ and $\vc{K}_Y:=(k_Y(Y_i,Y_j))_{i,j=1}^n $.  We need the centered kernel Gram matrices $\vc{G}_X=\vc{C}\vc{K}_X\vc{C}$ and $\vc{G}_Y=\vc{C}\vc{K}_Y\vc{C}$, where $ \vc{C} = \vc{I}_n -\frac{1}{n}\vc{D}_n$ with  $\vc{D}_n = \vc{1}_n\vc{1}^T_n$ and $\vc{1}_n$ is the vector with $n$ ones. The empirical estimate of Eq. (\ref{ckcca1}) is then given by
\begin{eqnarray}
\label{ckcca61}
\hat{\rho}=\max_{\substack{f_{X}\in \mc{H}_X,f_{Y}\in \mc{H}_Y \\ f_{X}\ne 0,\,f_{Y}\ne 0}}\frac{\widehat{\rm{Cov}}(f_X(X),f_Y(Y))}{[\widehat{\rm{Var}}(f_X(X))+\kappa\|f_X\|_{\mc{H}_X}]^{1/2}[\widehat{\rm{Var}}(f_Y(Y))+\kappa\|f_Y\|_{\mc{H}_Y}]^{1/2}}
\end{eqnarray}
where
\begin{align*}
& \widehat{\rm{Cov}}(f_X(X),f_Y(Y))
= \frac{1}{n} \vc{a}_X^T\vc{G}_X\vc{G}_Y \vc{a}_Y= \vc{a}_X^T\vc{G}_X\vc{W}\vc{G}_Y \vc{a}_Y , \\
& \widehat{\rm{Var}}( f_X(X))
=\frac{1}{n} \vc{a}_X^T\vc{G}_X^2 \vc{a}_X= \vc{a}_X^T\vc{G}_X \vc{W} \vc{G}_X \vc{a}_X, \,  \\ &\widehat{\rm{Var}}( f_Y(Y))=\frac{1}{n} \vc{a}_Y^T\vc{G}_Y^2 \vc{a}_Y= \vc{a}_Y^T\vc{G}_Y \vc{W}\vc{G}_Y\vc{a}_Y,
\end{align*}
and  $\vc{W}$ is a diagonal matrix with elements  $\frac{1}{n}$, and  $\vc{a}_{X}$ and $\vc{a}_{Y}$ are the   eigen-direction of  $X$ and $Y$, respectively. The regularized coefficient $\kappa > 0$. Solving the maximization problem in  Eq. (\ref{ckcca61})  is  analogous to solving the  following generalized eigenvalue problem:
\begin{align}
\label{ckcca71}
 \left[\vc{G}_Y\vc{W}\vc{G}_X  (\vc{G}_X \vc{W}\vc{G}_X+\kappa \vc{I})^{-\frac{1}{2}} \vc{G}_X\vc{W}\vc{G}_Y- \rho^2  (\vc{G}_Y \vc{W}\vc{G}_Y+\kappa \vc{I})\right] \vc{a}_{Y}=0
 \end{align}
\begin{align}
\label{ckcca72}
 \left[ \vc{G}_X\vc{W}\vc{G}_Y  (\vc{G}_Y \vc{W}\vc{G}_Y+\kappa \vc{I})^{-\frac{1}{2}} \vc{G}_Y\vc{W}\vc{G}_X - \rho^2  (\vc{G}_X \vc{W}\vc{G}_X+\kappa \vc{I})\right]\vc{a}_{X}=0
 \end{align}

It is easy to show  that the eigenvalues of  Eq. (\ref{ckcca71}) and Eq.   (\ref{ckcca72}) are  equal and that the eigenvectors for  any equation   can  be obtained from the other. The square roots of the eigenvalues of Eq.(\ref{ckcca71}) or  Eq. (\ref{ckcca72}) are the estimated kernel CC,  $\hat{\rho}$.   The $\hat{\rho}_j$ is the {\it j}th largest kernel CC and  the  {\it j}th kernel CVs are $\vc{a}_{X}^T\vc{G}_X$, and $\vc{a}_{Y}^T\vc{G}_Y$.

  Standard kernel CCA can be formulated using kernel CCO, which makes the robustness analysis easier.   As in \cite{Fukumizu-SCKCCA}, using the cross-covariance operator of (X,Y), $\Sigma_{XY}: \mc{H}_Y\to \mc{H}_X$  we can reformulate the optimization problem in Eq. (\ref{ckcca1})  as follows:
\begin{eqnarray}
\label{ckcca2}
\sup_{\substack{f_{X}\in \mc{H}_X, f_{Y}\in \mc{H}_Y \\ f_{X}\ne 0,\,f_{Y}\ne 0}}\langle f_X,\Sigma_{XY}f_Y\rangle_{\mc{H}_X}\qquad
\text{subject to}\qquad
\begin{cases}
	\langle f_X, \Sigma_{XX}f_X\rangle_{\mc{H}_X}=1,
\\
\langle f_Y, \Sigma_{YY}f_Y\rangle_{\mc{H}_Y}=1.
\end{cases}
\end{eqnarray}
 As  with linear CCA \citep{Anderson-03}, we can derive the solution of Eq. (\ref{ckcca2}) using the following   generalized eigenvalue problem.
\begin{eqnarray}
\label{ckcca3}
\begin{cases}
\Sigma_{XY}f_X - \rho\Sigma_{YY} f_Y = 0,
\\
\Sigma_{XY}f_Y - \rho\Sigma_{XX}f_X = 0.
\end{cases} \nonumber
\end{eqnarray}
 The  eigenfunctions of Eq. (\ref{ckcca3}) correspond to the largest eigenvalue,   which is  the solution to the kernel CCA problem.  After some simple calculations, we  reset the solution as
\begin{eqnarray}
\label{ckcca4}
\begin{cases}
(\Sigma_{XY} \Sigma_{YY}^{-1}\Sigma_{XY} - \rho^2 \Sigma_{XX})f_X = 0,\\
(\Sigma_{XY} \Sigma_{XX}^{-1}\Sigma_{XY}-\rho^2 \Sigma_{YY})f_Y = 0.
\end{cases}
\end{eqnarray}

It is known that the inverse of an operator may not exist. Even if it exists, it may not be continuous in general \citep{Fukumizu-SCKCCA}. We can derive kernel CC using the correlation operator
$\Sigma_{YY}^{- \frac{1}{2}}\Sigma_{YX} \Sigma_{XX}^{- \frac{1}{2}}$, even when $\Sigma_{XX}^{- \frac{1}{2}}$ and  $\Sigma_{YY}^{- \frac{1}{2}}$ are not proper operators. The potential danger  is that it might overfit, which is why introducing $\kappa$ as a regularization coefficient would be helpful.   Using the regularized coefficient $\kappa > 0$, the empirical estimators of Eq. (\ref{ckcca2}) and Eq. (\ref{ckcca4}) are
\begin{eqnarray}
\sup_{\substack{f_{X}\in \mc{H}_X,f_{Y}\in \mc{H}_X \\ f_{X}\ne 0,\,f_{Y}\ne 0}}\langle f_Y,\hat{\Sigma}_{YX}f_X\rangle_{\mc{H}_Y}\qquad
\text{subject to}\qquad
\begin{cases}
\langle f_X, (\hat{\Sigma}_{XX} + \kappa\vc{I}) f_X\rangle_{\mc{H}_X} = 1,
\\
\langle f_Y, (\hat{\Sigma}_{YY}+\kappa\vc{I})f_Y\rangle_{\mc{H}_Y} = 1,
\end{cases}
\label{ckcca73}
\end{eqnarray}
and
\begin{eqnarray}
\label{ckcca8}
\begin{cases}
(\hat{\Sigma}_{XY} (\hat{\Sigma}_{YY} + \kappa\vc{I})^{-1}\hat{\Sigma}_{XY} - \rho^2 (\hat{\Sigma}_{XX}+\kappa\vc{I}))f_X = 0,
\\
(\hat{\Sigma}_{YX} (\hat{\Sigma}_{XX} + \kappa\vc{I})^{-1}\hat{\Sigma}_{YX} - \rho^2 (\hat{\Sigma}_{YY}+\kappa\vc{I}))f_Y = 0,
\end{cases}
\end{eqnarray}
respectively.

   Now we calculate a finite rank operator  $ \mb{B}_{YX} = (\hat{\Sigma}_{{YY}} + \kappa \vc{I})^{- \frac{1}{2}} \hat{\Sigma}_{{YX}} (\hat{\Sigma}_{{XX}} + \kappa\vc{I})^{- \frac{1}{2}}$.  For $\kappa > 0$,  the square roots of the {\it j}-th   eigenvalue  of $\mb{B}_{YX}$ are the   {\it j}-th kernel CC, $\rho_j$.   The unit  eigenfunctions of   $\mb{B}_{YX}$  corresponding to the  {\it j}th eigenvalues  are $\hat{\nu}_{jX}\in \mc{H}_X$ and $\hat{\nu}_{j_Y}\in \mc{H}_Y$. The {\it j}th ($j= 1, 2, \cdots, n$)   kernel  CVs are
\[ \hat{f}_{jX}(X) = \langle \hat{f}_{jX}, \tilde{k}_X(\cdot, X)\rangle \,\rm{and}\,\hat{f}_{jY}(X) = \langle \hat{f}_{jY}, \tilde{k}_Y(\cdot, Y) \rangle\]
 where   $\hat{f}_{jX} =  (\hat{\Sigma}_{{XX}} + \kappa\vc{I})^{-\frac{1}{2}}\hat{\nu}_{jX}$ and   $\hat{f}_{jY} =  (\hat{\Sigma}_{{YY}} + \kappa\vc{I})^{-\frac{1}{2}}\hat{\nu}_{jY}.$

The generalized eigenvalue problem  in  Eq. (\ref{ckcca4})  can be formulated as a simple eigenvalue problem. Using the  {\it j}-th eigenfunction in  the first equation  of Eq. (\ref{ckcca4})  we have
\begin{eqnarray}
\label{ckcca5}
(\Sigma_{XX}^{- \frac{1}{2}} \Sigma_{XY} \Sigma_{YY}^{-1}\Sigma_{YX}\Sigma_{XX}^{- \frac{1}{2}} - \rho_j^2I) \Sigma_{XX}^{\frac{1}{2}}f_{jX}&=&0 \nonumber \\
\Rightarrow (\Sigma_{XX}^{-\frac{1}{2}} \Sigma_{XY} \Sigma_{YY}^{-1}\Sigma_{YX}\Sigma_{XX}^{- \frac{1}{2}} - \rho_j^2I)e_{jX}& = &0
\end{eqnarray}
where $e_{jX} = \Sigma_{XX}^{\frac{1}{2}}f_{jX}$.
\subsection{Influence function of the standard kernel canonical correlation analysis}
\label{sec:IFKCCA}
By using  the   IF   of  kernel PCA, linear PCA and   linear CCA,   we  can derive  the IF of kernel CCA (kernel CC and kernel CVs).
For simplicity, let us define $ \tilde{f}_X(X)=\langle f_X,  \tilde{k}_X (\cdot, X)$,  $\mb{L}_{jX}= \Sigma_{XX}^{- \frac{1}{2}}(\Sigma_{XX}^{- \frac{1}{2}} \Sigma_{XY} \Sigma_{YY}^{-1} \Sigma_{YX}\Sigma_{XX}^{- \frac{1}{2}}-\rho^2_j\vc{I})^{-1}\Sigma_{XX}^{- \frac{1}{2}}$, and  $\mb{L}_{jY}= \Sigma_{YY}^{- \frac{1}{2}}(\Sigma_{YY}^{- \frac{1}{2}} \Sigma_{YX} \Sigma_{XX}^{-1} \Sigma_{XY}\Sigma_{YY}^{- \frac{1}{2}}-\rho^2_j\vc{I})^{-1}\Sigma_{YY}^{- \frac{1}{2}}$.

\begin{thm}
\label{TIFKCCA}
 Given two sets of random variables $(X, Y)$ having the  distribution  $F_{XY}$ and the  { \it j}-th kernel CC ( $\rho_j$) and kernel  CVs ($f_{jX}(X)$ and $f_{jX}(Y)$), the  influence functions of  kernel CC  and kernel CVs  at $Z^\prime = (X^\prime, Y^\prime)$ are
\begin{multline}
\rm{IF} (Z^\prime, \rho_j^2)= - \rho_j^2 \tilde{f}_{jX}^2(X^\prime) + 2 \rho_j \tilde{f}_{jX}(X^\prime) \tilde{f}_{jY}(Y^\prime)  - \rho_j^2 \tilde{f}_{jY}^2(Y^\prime), \nonumber\\
\rm{IF} (\cdot, Z^\prime, f_{jX}) = -\rho_j [\tilde{f}_{jY}(Y^\prime) - \rho_j \tilde{f}_{jX}(X^\prime)]\mb{L}_{jX} \tilde{k}_X (\cdot, X^\prime) -  [\tilde{f}_{jX}(X^\prime)  - \rho_j \tilde{f}_{jY}(Y^\prime)]\mb{L}_{jX}\Sigma_{XY}\Sigma^{-1}_{YY} \tilde{k}_Y(\cdot,  Y^\prime)\nonumber\\\qquad\qquad\qquad\qquad\qquad\qquad\qquad\qquad\qquad\qquad\qquad\qquad+\frac{1}{2}[1- \tilde{f}^2_{jX}(X^\prime)]f_{jX}, \nonumber\\
 \rm{IF} (\cdot, Z^\prime, f_{jY})
= -\rho_j [\tilde{f}_{jX}(X^\prime) - \rho_j \tilde{f}_{jY}(Y^\prime)]\mb{L}_{jY} \tilde{k}_Y(\cdot, Y^\prime)-  [\tilde{f}_{jY}(Y^\prime)- \rho_j \tilde{f}_{jX}(X^\prime)]\mb{L}_{jY}\Sigma_{YX}\Sigma^{-1}_{XX} \tilde{k}_Y(\cdot, Y^\prime) \nonumber\\ +\frac{1}{2}[1- \tilde{f}^2_{jY}(Y^\prime)]f_{jY}. \nonumber
\end{multline}
\end{thm}
The above theorem  has been  proved on the basis of previously established ones, such as the IF of  linear  PCA \citep{Tanaka-88, Tanaka-89}, the IF of linear CCA \citep{Romanazii-92},  and the IF of  kernel PCA, respectively. To do this,  we  convert the generalized eigenvalue problem of kernel CCA  into a simple eigenvalue problem. First, we need to find the IF of  $ \Sigma_{XX}^{- \frac{1}{2}} \Sigma_{XY} \Sigma_{YY}^{-1}\Sigma_{YX}\Sigma_{XX}^{- \frac{1}{2}}$, henceforth the IF of $\Sigma_{YY}^{-1},  \Sigma_{XX}^{\frac{1}{2}}$ and $\Sigma_{XY}$.

 Using the above  result,  we can establish some properties of kernel CCA: robustness, asymptotic consistency and its standard error. In addition, we  are able to identify the outliers based on the influence of the data. All notations and proof are explained in the appendix.

  The IF of inverse covariance operator exists  only for the finite dimensional RKHS. For infinite dimensional RKHS,  we can find the IF of $\Sigma_{XX}^{- \frac{1}{2}}$  by introducing  a regularization term as follows
\begin{multline}
\rm{IF}(\cdot, X^\prime,
(\Sigma_{XX} + \kappa\vc{I})^{- \frac{1}{2}}) =\nonumber\\ \frac{1}{2} [(\Sigma_{XX}+\kappa\vc{I})^{- \frac{1}{2}}- (\Sigma_{XX}+\kappa\vc{I})^{- \frac{1}{2}}\tilde{k}_X(\cdot, X^\prime)\otimes  \tilde{k}_X(\cdot, X^\prime)(\Sigma_{XX} + \kappa\vc{I})^{- \frac{1}{2}}], \nonumber
\end{multline}
which gives the empirical estimator.

Let $(X_i, Y_i)_{i=1}^n$ be a sample from the  empirical joint distribution $F_{nXY}$. The EIF (IF based on empirical distribution) of kernel CC and  kernel CVs  at $(X^\prime, Y^\prime)$ for all points $ (X_i, Y_i)$ are $\rm{EIF} (X_i, Y_i, X^\prime, Y^\prime, \rho_j^2) = \widehat{\rm{IF}} ( X^\prime, Y^\prime,  \hat{\rho}_j^2)$, $\rm{EIF} (X_i, Y _i,  X^\prime, Y^\prime, f_{jX}) = \widehat{\rm{IF}} (\cdot, X^\prime, Y^\prime, f_{jX})$, and $\rm{EIF} (X_i, Y_i,   X^\prime, Y^\prime, f_{jY}) = \widehat{\rm{IF}} (\cdot,  X^\prime, Y^\prime, \widehat{f}_{jY})$, respectively.

\subsection{Robust kernel canonical correlation analysis}
\label{sec:RKCCA}
In this section, we propose  a {\bf robust kernel CCA} method based on the  robust kernel CO and the  robust kernel CCO. While many robust linear CCA  methods have been proposed to show  that linear CCA methods cannot fit the bulk of the data  and have  points deviating from the original pattern  for further investment \citep{Adrover-15, Ashad-10}, there are no  well-founded robust methods of  kernel CCA. The  standard kernel CCA  considers the same weights for each data point, $\frac{1}{n}$,  to estimate kernel CO and kernel CCO, which is the solution of an empirical risk optimization problem when using the quadratic loss function. It is known that the least square loss function is not a robust loss function. Instead, we can solve  an empirical risk optimization problem using the robust least square loss function where  the weights are  determined based on  KIRWLS. We need robust centered kernel Gram matrices of $X$ and $Y$ data.  The  centered  robust kernel Gram matrix of $X$ is,   $\vc{G}_{RX}=\vc{C_{R}}\vc{K}_X\vc{C_{R}}$  where $ \vc{C}_R = \vc{I}_n - \vc{1}_n\vc{w}^T_n$, $\vc{1}_n$ is the vector with $n$ ones and $\vc{w}$ is a weight vector of robust kernel ME, $\mc{M}_X$. Similarly,  we can calculate  $\vc{G}_{RY}$ for $Y$.  After getting robust kernel CO and kernel CCO, they are used in standard kernel CCA, which we call {\bf robust kernel CCA}.  The empirical estimate of Eq. (\ref{ckcca1}) is then given by
\begin{eqnarray}
\label{ckcca6}
\hat{\rho}_{rkcc}=\max_{\substack{g_{X}\in \mc{H}_X,g_{Y}\in \mc{H}_Y \\ g_{X}\ne 0,\,g_{Y}\ne 0}}\frac{\widehat{\rm{Cov}_R}(g_X(X),g_Y(Y))}{[\widehat{\rm{Var}_R}(g_X(X))+\kappa\|g_X\|_{\mc{H}_X}]^{1/2}[\widehat{\rm{Var}_R}(g_Y(Y))+\kappa\|g_Y\|_{\mc{H}_Y}]^{1/2}} \nonumber
\end{eqnarray}
 with $g_X\in {\mc{H}}_X$,  $g_Y\in {\mc{H}}_Y$ and
\begin{align*}
& \widehat{\rm{Cov}_R}(g_X(X),g_Y(Y))
=  \vc{b}_X^T\vc{G}_{RX}\vc{W}_{XY}\vc{G}_{RY} \vc{b}_Y , \\
& \widehat{\rm{Var}_R}( g_X(X))
= \vc{b}_X^T\vc{G}_{RX} \vc{W}_{XX} \vc{G}_{RX} \vc{b}_X, \,  \\ &\widehat{\rm{Var}_R}( g_Y(Y))= \vc{b}_Y^T\vc{G}_{RY} \vc{W}_{YY} \vc{G}_{RY}\vc{b}_Y,
\end{align*}
 where $\vc{W}_{XY}$,  $\vc{W}_{XX}$, and $\vc{W}_{YY}$ are   diagonal matrices with elements corresponding to the  weights of  robust kernel CCO, and kernel COs, respectively.  Also  $\vc{b}_{X}$ and $\vc{b}_{Y}$ are the   eigen-direction  of   $X$ and $Y$, respectively. As in  Eq. (\ref{ckcca71}), we can   solve the  maximization problem of Eq. (\ref{ckcca6}) as an  eigenvalue problem.
Let  $\Sigma_{RXY}$, $\Sigma_{RXX}$, and $\Sigma_{RYY}$ be the  robust kernel CCO, robust kernel CO of $X$, and   robust kernel CO of $Y$, respectively. Like standard kernel CCA, the  robust empirical estimators of Eq. (\ref{ckcca2}) and Eq. (\ref{ckcca4}) are
\begin{eqnarray}
\sup_{\substack{f_{X}\in \mc{H}_X,f_{Y}\in \mc{H}_X \\ f_{X}\ne 0,\,f_{Y}\ne 0}}\langle f_Y,\hat{\Sigma}_{RXY}f_X\rangle_{\mc{H}_Y}\qquad
\text{subject to}\qquad
\begin{cases}
\langle f_X, (\hat{\Sigma}_{RXX} + \kappa\vc{I}) f_X\rangle_{\mc{H}_X} = 1,
\\
\langle f_Y, (\hat{\Sigma}_{RYY}+\kappa\vc{I})f_Y\rangle_{\mc{H}_Y} = 1,
\end{cases}
\label{rkcca73}
\end{eqnarray}
and
\begin{eqnarray}
\label{rkcca8}
\begin{cases}
(\hat{\Sigma}_{RXY} (\hat{\Sigma}_{RYY} + \kappa\vc{I})^{-1}\hat{\Sigma}_{RXY} - \rho^2 (\hat{\Sigma}_{RXX}+\kappa\vc{I}))f_X = 0,
\\
(\hat{\Sigma}_{RXY} (\hat{\Sigma}_{RXX} + \kappa\vc{I})^{-1}\hat{\Sigma}_{RXY} - \rho^2 (\hat{\Sigma}_{RYY}+\kappa\vc{I}))f_Y = 0,
\end{cases}
\end{eqnarray}
respectively.
This method  is designed for  contaminated data, and  the principles we describe also apply to the  kernel methods, which must deal with the issue of kernel CO and kernel CCO.

\section{An R Package, RKUM}

We implemented an R package RKUM  to facilitate inference on the robust kernel unsupervised methods \citep{Alam-18a, Alam-18b}.
\begin{example}
mdbw <-
function(X)
  {
   n <- dim(X)[1]
   ab <- X%*%t(X) 
   aa <- as.matrix(diag(ab))
   Dx <-matrix(rep(aa,each=n), ncol=n, byrow=TRUE) +  matrix(rep(aa,each=n),nrow=n) - 2*ab  
   Dx <- Dx-diag(diag(Dx))
   dx <- matrix(Dx, n*n,1)
   s <- sqrt(median(dx[dx!=0]))
  return (s)
 }
\end{example}

\begin{example}
gkm <-
function(X)
   {
     n <- dim(X)[1]
     SGX <-  mdbw(X)
     sx2  <- 2*SGX*SGX
     ab <- X%*%t(X) 
     aa <- as.matrix(diag(ab))
     D <- matrix(rep(aa,each=n), ncol=n, byrow=TRUE) 
     xx <-pmax(D + t(D) - 2*ab,  mat.or.vec(n, n))
     K <- exp(-xx/sx2)  
     return(K)
   }

\end{example}

\begin{verbatim}
hulfun <- function(x) {
  n <- length(x)
  a <- median(x)
  y <- rep(0, n)
  for (i in 1:n) {
    if (x[i] <= a) {
      y[i] <- x[i]^2 / 2
    } else {
      y[i] <- a * x[i] - a^2 / 2
    }
  }
  return(y)
}
\end{verbatim}

\begin{example}
hudr <-
function(x) 
   {
    n<-length(x)
    a <- median(x)
    y <- rep(0, n)
    for (i in 1: n)
    {
       if( x[i] <= (a))
       {
        y[i] <-1
        }
        else
        y[i] <- a/x[i]
    }
  return(y)
 }
\end{example}

\begin{example}
hulofun <-
function(x)
  {
   y<- hulfun(x)
   obj<-mean(y)
  return(obj)
  }

\end{example}

\begin{example}
rkcm <-
function(X, lossfu ="Huber", kernel="rbfdot")     
   {
    n <-dim(X)[1]
  if (lossfu=="square")
    {
      if(kernel=="linear")
     {
     K <- lkm(X)
     Id <- diag(1, nrow = n)
     Id1 <- matrix(1, nrow = n, ncol = n)
     H <- Id - Id1/n
     K.robust.M <- tcrossprod(H%*%K, H)
      }
   if(kernel=="rbfdot")
     {
     K <- gkm(X)
    Id <- diag(1, nrow = n)
    Id1 <- matrix(1, nrow = n, ncol = n)
    H <- Id - Id1/n
    K.robust.M <- tcrossprod(H%*%K, H)
     }
   if(kernel=="ibs")
     {
     K <- ibskm(X)
    Id <- diag(1, nrow = n)
    Id1 <- matrix(1, nrow = n, ncol = n)
    H <- Id - Id1/n
    K.robust.M <- tcrossprod(H%*%K, H)
    }
  }
   if(lossfu=="Hampel")
   {
    if(kernel=="linear")
     {
     K <- lkm(X)
     W <- rep(1/n, n)
    aa<- sqrt(abs(diag(K)-2*as.vector(crossprod(W,K))+ as.vector(crossprod(W,K%*%W))))
    for (k in 1:100)
    {
       Obj.old<- halofun(aa)
        HV <-hadr(aa)
        THV <- sum(HV)
        W <-HV/THV
  R_mean_E <-as.vector(crossprod(W,K))
  aa<- sqrt(abs(diag(K)-2*as.vector(crossprod(W,K))+ as.vector(crossprod(W,K%*%W))))
        Obj.new<- halofun(aa)
        Stop <- abs(Obj.old-Obj.new)/Obj.old
        if( Stop < 0.1^8)
          {
            break
          }
     }
    ee<- rep(1, n)
    H <- diag(1, n)- as.vector(tcrossprod(ee,W))
    K.robust.M <-tcrossprod(H%*%K,H)
     }
    if(kernel=="rbfdot")
     {
     K <- gkm(X) 
   W <- rep(1/n, n)
    aa<- sqrt(abs(diag(K)-2*as.vector(crossprod(W,K))+ as.vector(crossprod(W,K%*%W))))
    for (k in 1:100)
    {
       Obj.old<- halofun(aa)
        HV <-hadr(aa)
        THV <- sum(HV)
        W <-HV/THV
  R_mean_E <-as.vector(crossprod(W,K))
  aa<- sqrt(abs(diag(K)-2*as.vector(crossprod(W,K))+ as.vector(crossprod(W,K%*%W))))
        Obj.new<- halofun(aa)
        Stop <- abs(Obj.old-Obj.new)/Obj.old
        if( Stop < 0.1^8)
          {
            break
          }
     }
    ee<- rep(1, n)
    H <- diag(1, n)- as.vector(tcrossprod(ee,W))
    K.robust.M <-tcrossprod(H%*%K,H)
    }
   if(kernel=="ibs")
     {
     K <- ibskm(X)
    W <- rep(1/n, n)
    aa<- sqrt(abs(diag(K)-2*as.vector(crossprod(W,K))+ as.vector(crossprod(W,K%*%W))))
    for (k in 1:100)
    {
       Obj.old<- halofun(aa)
        HV <-hadr(aa)
        THV <- sum(HV)
        W <-HV/THV
  R_mean_E <-as.vector(crossprod(W,K))
  aa<- sqrt(abs(diag(K)-2*as.vector(crossprod(W,K))+ as.vector(crossprod(W,K%*%W))))
        Obj.new<- halofun(aa)
        Stop <- abs(Obj.old-Obj.new)/Obj.old
        if( Stop < 0.1^8)
          {
            break
          }
     }
    ee<- rep(1, n)
    H <- diag(1, n)- as.vector(tcrossprod(ee,W))
    K.robust.M <-tcrossprod(H%*%K,H)
    }
  }
  if(lossfu=="Huber")
    {
    if (kernel=="linear")
     {
     K <- lkm(X)
     W <- rep(1/n, n)
    aa<- sqrt(diag(K)-2*as.vector(crossprod(W,K))+ as.vector(crossprod(W,K%*%W)))
    for (k in 1:100)
    {
        Obj.old<- hulofun(aa)
        HV <- hudr(aa)
        THV <- sum(HV)
        W <-HV/THV
      R_mean_E <-as.vector(crossprod(W,K))#as.vector(t(W)%*%K)
      aa<- sqrt(diag(K)-2*as.vector(crossprod(W,K))+ as.vector(crossprod(W,K%*%W)))
        Obj.new<- hulofun(aa)
        Stop <- abs(Obj.old-Obj.new)/Obj.old
        if( Stop < 0.1^8)
          {
            break
          }
     }
    ee<- rep(1, n)
    H <- diag(1, n)- as.vector(tcrossprod(ee,W))
    K.robust.M <-tcrossprod(H%*%K,H)
     }
    if(kernel=="rbfdot")
     {
     K <- gkm(X)
     W <- rep(1/n, n)
    aa<- sqrt(diag(K)-2*as.vector(crossprod(W,K))+ as.vector(crossprod(W,K%*%W)))
    for (k in 1:100)
    {
        Obj.old<- hulofun(aa)
        HV <- hudr(aa)
        THV <- sum(HV)
        W <-HV/THV
      R_mean_E <-as.vector(crossprod(W,K))#as.vector(t(W)%*%K)
      aa<- sqrt(diag(K)-2*as.vector(crossprod(W,K))+ as.vector(crossprod(W,K%*%W)))
        Obj.new<- hulofun(aa)
        Stop <- abs(Obj.old-Obj.new)/Obj.old
        if( Stop < 0.1^8)
          {
            break
          }
     }
    ee<- rep(1, n)
    H <- diag(1, n)- as.vector(tcrossprod(ee,W))
    K.robust.M <-tcrossprod(H%*%K,H)
    }
   if(kernel=="ibs")
     {
     K <- ibskm(X)
      W <- rep(1/n, n)
    aa<- sqrt(diag(K)-2*as.vector(crossprod(W,K))+ as.vector(crossprod(W,K%*%W)))
    for (k in 1:100)
    {
        Obj.old<- hulofun(aa)
        HV <- hudr(aa)
        THV <- sum(HV)
        W <-HV/THV
      R_mean_E <-as.vector(crossprod(W,K))#as.vector(t(W)%*%K)
      aa<- sqrt(diag(K)-2*as.vector(crossprod(W,K))+ as.vector(crossprod(W,K%*%W)))
        Obj.new<- hulofun(aa)
        Stop <- abs(Obj.old-Obj.new)/Obj.old
        if( Stop < 0.1^8)
          {
            break
          }
     }
    ee<- rep(1, n)
    H <- diag(1, n)- as.vector(tcrossprod(ee,W))
    K.robust.M <-tcrossprod(H%*%K,H)
   }
  }
   K.robust.M <- (K.robust.M + t(K.robust.M))/2
  return(list(rkcm=K.robust.M)) 
 }

\end{example}

\begin{example}
rkcco <-
function(X, Y, lossfu ="Huber", kernel="rbfdot",  gamma= 0.00001) 
 {                      
   n <- nrow(X)
   RTX <- rkcm(X, lossfu, kernel)
   RTY <- rkcm(Y, lossfu, kernel)
   CKX <- RTX$rkcm
   CKY <- RTY$rkcm
   if (lossfu=="square")
     {
      W <- rep(1/n, n)
      RXX1 <- (CKX+diag(rep(gamma,n)))%*%diag(W)%*%CKX 
      RYY1 <- (CKY+diag(rep(gamma,n)))%*%diag(W)%*%CKY 
      RXY1 <- (CKX+diag(rep(gamma,n)))%*%diag(W)%*%CKY 
     }
  if (lossfu=="Hampel")
    {
    W <- rep(1/n, n) 
    error <- sqrt(abs(diag(CKX)*diag(CKY)-2*as.vector(crossprod(W,(CKX*CKY)))+as.vector(crossprod(W,  (CKX*CKY)%*%W))))
    for (k in 1:100)
    {
      Obj.old <- halofun(error)
      HV <- hadr(error)
     THV <- sum(HV)
      W <- HV/THV
      R_SecM <-  tcrossprod(CKX%*%diag(W),CKY)
      error <- sqrt(abs(diag(CKX)*diag(CKY)-2*as.vector(crossprod(W,(CKX*CKY)))+as.vector(crossprod(W,(CKX*CKY)%*%W))))
    Obj.new <-  halofun(error)
      Stop <- abs(Obj.old-Obj.new)/Obj.old
      if( Stop < 0.1^8)
        {
          break
        }
    }
     RXX1 <- (CKX+diag(rep(gamma,n)))%*%diag(W)%*%CKX 
     RYY1 <- (CKY+diag(rep(gamma,n)))%*%diag(W)%*%CKY 
     RXY1 <- (CKX+diag(rep(gamma,n)))%*%diag(W)%*%CKY 
   }
  if (lossfu=="Huber")
   {
   W <- rep(1/n, n) 
   error <- sqrt(abs(diag(CKX)*diag(CKY)-2*as.vector(crossprod(W,(CKX*CKY)))+as.vector(crossprod(W,  (CKX*CKY)%*%W))))
   for (k in 1:100)
    {
      Obj.old <- hulofun(error)
      HV <- hudr(error)
    THV <- sum(HV)
      W <- HV/THV
      R_SecM <-  tcrossprod(CKX%*%diag(W),CKY)
      error <- sqrt(abs(diag(CKX)*diag(CKY)-2*as.vector(crossprod(W,(CKX*CKY)))+as.vector(crossprod(W,(CKX*CKY)%*%W))))
    Obj.new <-  hulofun(error)
      Stop <- abs(Obj.old-Obj.new)/Obj.old
       if( Stop < 0.1^8)
        {
          break
        }
     }
     RXX1 <- (CKX+diag(rep(gamma,n)))%*%diag(W)%*%CKX 
     RYY1 <- (CKY+diag(rep(gamma,n)))%*%diag(W)%*%CKY 
     RXY1 <- (CKX+diag(rep(gamma,n)))%*%diag(W)%*%CKY 
    }
   RXX <- (RXX1 + t(RXX1))/2
   RYY <- (RYY1 + t(RYY1))/2 
   RXY <-RXY1  
  return(list(rkcmx = CKX, rkcmy = CKY, rkcooxx = RXX, rkcooyy=RYY, rkcooxy =RXY)) 
 }
\end{example}

\begin{example}
rkcca <-
function (X,  Y, lossfu ="Huber", kernel="rbfdot", gamma=0.00001, ncomps=10) 
    {
    RXY <- rkcco(X, Y, lossfu, kernel, gamma)    
    CKX <- RXY$rkcmx
    CKY <- RXY$rkcmy
    KXX <- RXY$rkcooxx
    KYY <- RXY$rkcooyy
    KXY <- RXY$rkcooxy
    n <- dim(KXX)[1]
    m <- 2
    AB <- matrix(0, n *2, 2*n)
    AB[0:n,0:n ] <- 0
    AB[0:n,(n + 1):(2 * n)] <- (KXY)
    AB[(n + 1):(2 * n),0:n ] <- t(KXY)
    AB[(n + 1):(2 * n),(n + 1):(2 * n) ] <-0
    CD <- matrix(0, n * m, n * m)
    CD[1:n, 1:n] <- (KXX+diag(rep(1e-6,n))) 
    CD[(n + 1):(2 * n), (n + 1):(2 * n)] <- (KYY+diag(rep(1e-6,n)))
    CD <- (CD + t(CD))/2
    ei <- gmedc(AB, CD)
    kcor<- abs(as.double(ei$gvalues[1:ncomps]))
    xcoef <- matrix(as.double(ei$gvectors[1:n, 1:ncomps]),n)
    ycoef<- matrix(as.double(ei$gvectors[(n + 1):(2 * n), 
        1:ncomps]), n)
    CVX <- CKX%*%xcoef
    CVY <- CKY%*%ycoef
     return(list(rkcor = kcor, rxcoef = xcoef, rycoef = ycoef,  rxcv= CVX, rycv=CVY))
 }

\end{example}

\begin{example}
snpfmridata <-
function(n=300, gamma=0.00001, ncomps=10, jth=1)
   {
      sig1 <- 0.5 
sig2 <- 1 
p1 <- 100
p2 <- 100
      pt <- 100 
a <- rep(0, pt)
      a[1:p1] <- ranuf(p1)
b <- rep(0,pt)
b[1:p2] <- ranuf(p2) #using rndu.r file
maf  <-  runif(pt,0.2, 0.4) # (0,1)/5+0.2  # form uiform distribution
mu <- rnorm(n, 0, 1)
      mu <- sign(mu)*(abs(mu)+0.1);# normal distribution add some samll term 0.1
mu <-mu*sig1 
      ind <- rep(F,pt)        #logical(pt) # logical zeors
      ind <- as.integer(ind)
ind[1:p2] <- 1 # take 1 1: p2  matrix( rnorm(N*M,mean=0,sd=1), N, M) 
m1 <- mu%*%t(a)+matrix(rnorm(n*pt, 0, 1), n, pt)*sig1 #fmri
m2 <- mu%*%t(b)+matrix(rnorm(n*pt, 0, 1), n, pt)*sig2 # SNP      
 for ( i in 1: pt)
 {
    p <- maf[i]
     bias <- log(1/p-1)
    if (ind[i]==TRUE)
  {
    gt <- m2[,i]
    gt <- 1/(1+exp(-gt+bias)) # for correlated variables case
        }
     else
     gt <- 1/(1+exp(bias))      #For no correlated variable #control
     temp <- runif(n,0,1)
  m2[,i] <- gt > temp
   temp <- runif(1, 0,1)
    m2[,i] <- m2[,i]+(gt > temp)
}
  E3X <- m1
  E3Y <- m2
n1 <- (0.05*n) 
      ss <- sample(1:n, n1)
 m1 <-mu[ss]%*%t(a)+matrix(rnorm(n1*pt, 0, 1), n1,pt)*10 #fmri
     m2 <- mu[ss]%*%t(b)+matrix(rnorm(n1*pt, 0, 1), n1,pt)*20
       for ( i in 1: pt)
 {
    p <- maf[i]
     bias <- log(1/p-1)
    if (ind[i]==TRUE)
  {
    gt <- m2[,i]
    gt <- 1/(1+exp(-gt+bias)) # for correlated variables
        }
     else
     gt <- 1/(1+exp(bias))      #For no correlated variable
     temp <- runif(n1,0,1)
  m2[,i] <- gt > temp
   temp <- runif(1, 0,1)
    m2[,i] <- m2[,i]+(gt > temp)
}

E3Xot <- rbind(E3X[-ss,],m1)
  E3Yot <- rbind(E3Y[-ss,],m2)
      CCAID <- ifcca(E3X,E3Y, gamma, ncomps, jth) 
      CCACD <- ifcca(E3Xot,E3Yot, gamma, ncomps, jth)  
      KCCAID <- ifrkcca(E3X,E3Y, "square", "linear", gamma, ncomps, jth)  
      KCCACD <- ifrkcca(E3Xot,E3Yot, "square", "linear", gamma, ncomps, jth)      
      HARKCCAID <- ifrkcca(E3X, E3Y, "Hampel", "linear", gamma, ncomps, jth) 
      HARKCCACD <- ifrkcca(E3Xot,E3Yot, "Hampel", "linear",gamma, ncomps, jth)       
      HURKCCAID <- ifrkcca(E3X,E3Y, "Huber", "linear", gamma, ncomps, jth) 
      HURKCCACD <- ifrkcca(E3Xot, E3Yot, "Huber", "linear",gamma, ncomps, jth) 
      CCAIFID <-  CCAID$iflccor 
      CCAIFCD <- CCACD$iflccor
      KCCAIFID <- KCCAID$ifrkcor 
      KCCAIFCD <- KCCACD$ifrkcor      
      HAKCCAIFID <- HARKCCAID$ifrkcor 
      HAKCCAIFCD <- HARKCCACD$ifrkcor 
      HUKCCAIFID <- HURKCCAID$ifrkcor 
      HUKCCAIFCD <- HURKCCAID$ifrkcor       
   {                           
      par(mfrow=c(4,2))
par(mar=c(5.1,5.1,4.1,2.1))
      plot(CCAIFID, pch=21:22,main=list("Ideal Data",cex =2, font=15), sub=list("Linear CCA",cex =2, font=15),ylab="",xlab=list(" ",cex =2, font=15),
     type="o",lwd=3,cex.axis=2,cex.lab=2,axes=FALSE)
     axis(2, las=0,cex.axis=2,cex.lab=2)
      plot(CCAIFCD, pch=21:22, main=list("Contaminated Data",cex =2, font=15),sub=list("Linear CCA",cex =2, font=15), ylab="",xlab=list(" ",cex =2, font=15),
     type="o",lwd=3,cex.axis=2,cex.lab=2,axes=FALSE)
     axis(2, las=0,cex.axis=2,cex.lab=2)
      plot(KCCAIFID, pch=21:22, main=list("",cex =2, font=15),sub=list("Kernel CCA",cex =2, font=15),ylab="",xlab=list(" ",cex =2, font=15),
     type="o",lwd=3,cex.axis=2,cex.lab=2,axes=FALSE)
     axis(2, las=0,cex.axis=2,cex.lab=2)
      plot(KCCAIFCD, pch=21:22, main=list("",cex =2, font=15),sub=list(" Kernel CCA",cex =2, font=15),ylab="",xlab=list(" ",cex =2, font=15),
     type="o",lwd=3,cex.axis=2,cex.lab=2,axes=FALSE)
     axis(2, las=0,cex.axis=2,cex.lab=2)
      plot(HAKCCAIFID, pch=21:22, main=list("",cex =2, font=15), sub=list(" Hampel's  robust kernel CCA",cex =2, font=15),ylab="",xlab=list(" ",cex =2, font=15),
     type="o",lwd=3,cex.axis=2,cex.lab=2,axes=FALSE)
     axis(2, las=0,cex.axis=2,cex.lab=2)
      plot(HAKCCAIFCD, pch=21:22, main=list("",cex =2, font=15),sub=list(" Hampel's robust kernel CCA",cex =2, font=15), ylab="",xlab=list(" ",cex =2, font=15),
     type="o",lwd=3,cex.axis=2,cex.lab=2,axes=FALSE)
     axis(2, las=0,cex.axis=2,cex.lab=2)
      plot(HUKCCAIFID, pch=21:22, main=list("",cex =2, font=15),sub=list(" Huber's robust kernel CCA",cex =2, font=15),ylab="",xlab=list(" ",cex =2, font=15),
     type="o",lwd=3,cex.axis=2,cex.lab=2,axes=FALSE)
     axis(2, las=0,cex.axis=2,cex.lab=2)
      plot(HUKCCAIFCD, pch=21:22, main=list("",cex =2, font=15),sub=list(" Huber's robust kernel CCA",cex =2, font=15),ylab="",xlab=list(" ",cex =2, font=15),
     type="o",lwd=3,cex.axis=2,cex.lab=2,axes=FALSE)
     axis(2, las=0,cex.axis=2,cex.lab=2)
     }
 return(list(IFCCAID = CCAIFID,  IFCCACD = CCAIFCD,  IFKCCAID = KCCAIFID,  IFKCCACD = KCCAIFCD, IFHAKCCAID = HAKCCAIFID,  IFHAKCCACD = HAKCCAIFCD, IFHUKCCAID = HUKCCAIFID,  IFHUKCCACD = HUKCCAIFCD))
 }
\end{example}

\begin{example}
snpfmrimth3D <-
function(n=500, gamma=0.00001, ncoms=1)
     {
      sig1 <- 0.5 #signal level
      sig2 <- 1 #noise level
      sig3 <- 0.1 #noise level
    p1 <- 100
    p2 <- 100
      p3 <- 100
      pt <- 100 # dimension for correlated voxels,correlated SNPs and data.
     a <- rep(0, pt)
      a[1:p1] <-  ranuf(p1)
     b <- rep(0,pt)
    b[1:p2] <-  ranuf(p2) #using rndu.r file
      maf  <-  runif(pt,0.2, 0.4) # (0,1)/5+0.2  # form uiform distribution
     mu <- rnorm(n, 0, 1)
      mu <- sign(mu)*(abs(mu)+0.1);# normal distribution add some samll term 0.1
    mu <-mu*sig1 # Common feature
      ind <- rep(F,pt)        #logical(pt) # logical zeors
      ind <- as.integer(ind)
    ind[1:p2] <- 1 # take 1 1: p2  matrix( rnorm(N*M,mean=0,sd=1), N, M) 
    m1 <- mu%*%t(a)+matrix(rnorm(n*pt, 0, 1), n,pt)*sig1 #fmri
    m2 <- mu%*%t(b)+matrix(rnorm(n*pt, 0, 1), n,pt)*sig2 # SNP  
         # Methlation 
      prop <- c(0.20,0.30,0.27,0.23)
      effect <- 2.5
      n.sample=n
      cluster.sample.prop = c(0.30,0.30,0.40)
      n.cluster <- length(cluster.sample.prop)
      delta.methyl=effect
      p.DMP=0.2
      sigma.methyl=NULL
      DMP <- sapply(1:n.cluster, function(x) rbinom(p3, 1, prob = p.DMP))
      d <- lapply(1:n.cluster,function(i) {
  effect <- DMP[,i]*delta.methyl
   mvnod(n=cluster.sample.prop[i]*n.sample, mu=effect, Sigma=diag(1, p3,p3))})
      sim.methyl <- do.call(rbind,d)
      m3 <- rlogit(sim.methyl) + matrix(rnorm(n*p3, 0, 1), n, p3)*sig3 
     #tO CONSTRACT Normal distribution o binomial distribution
   for ( i in 1: pt)
   {
    p <- maf[i]
     bias <- log(1/p-1)
    if (ind[i]==TRUE)
   {
    gt <- m2[,i]
    gt <- 1/(1+exp(-gt+bias)) # for correlated variables
        }
     else
     gt <- 1/(1+exp(bias))      #For no correlated variable
     temp <- runif(n,0,1)
  m2[,i] <- gt > temp
   temp <- runif(1, 0,1)
    m2[,i] <- m2[,i]+(gt > temp)
  }
  EX <- m1
    EY <- m2
      EZ <- m3
n1 <- (0.05*n) 
      ss <- sample(1:n, n1)
    m1 <-mu[ss]%*%t(a)+matrix(rnorm(n1*pt, 0, 1), n1,pt)*1.5 #fmri
     m2 <- mu[ss]%*%t(b)+matrix(rnorm(n1*pt, 0, 1), n1,pt)*2
     for ( i in 1: pt)
 {
    p <- maf[i]
     bias <- log(1/p-1)
    if (ind[i]==TRUE)
  {
    gt <- m2[,i]
    gt <- 1/(1+exp(-gt+bias)) # for correlated variables
        }
     else
     gt <- 1/(1+exp(bias))      #For no correlated variable
     temp <- runif(n1,0,1)
  m2[,i] <- gt > temp
   temp <- runif(1, 0,1)
    m2[,i] <- m2[,i]+(gt > temp)
}
 EXout <- rbind(EX[-ss,],m1)
EYout <- rbind(EY[-ss,],m2)
      EZout <- rbind(EZ[-ss,], rlogit(sim.methyl[ss,])+ matrix(rnorm(n1*p3, 0, 1), n1, p3)*3)    
      ReIF1 <- ifmkcca(EX, EY,  EZ, "linear", 0.00001, 1)
      ReIFout1 <- ifmkcca(EXout, EYout, EZout, "linear", 0.00001, 1)
     ReIF <-  ReIF1$ifmkcor
      ReIFout <-  ReIFout1$ifmkcor
       {
   par(mfrow=c(1,2))
   par(mar=c(5.1,5.1,4.1,2.1))
      plot(ReIF, pch=21:22, main=list("Ideal data",cex =2, font=15),sub=list("Multiple kernel CCA",cex =2, font=15), ylab="", xlab=list(" ",cex =2, font=15),
     type="o",lwd=3,cex.axis=2,cex.lab=2,axes=FALSE)
     axis(2, las=0,cex.axis=2,cex.lab=2)
      plot( ReIFout, pch=21:22, main=list("Contaminated data", cex =2, font=15), sub=list(" Multiple kernel CCA",cex =2, font=15), ylab="",xlab=list(" ",cex =2, font=15),
     type="o",lwd=3,cex.axis=2,cex.lab=2,axes=FALSE)
     axis(2, las=0,cex.axis=2,cex.lab=2)
      }
 return(list(IFim=  ReIF, IFcm =  ReIFout))
 }
\end{example}
\begin{figure}[htbp]
  \centering
  \includegraphics[width=\linewidth, height=0.9\textheight, keepaspectratio]{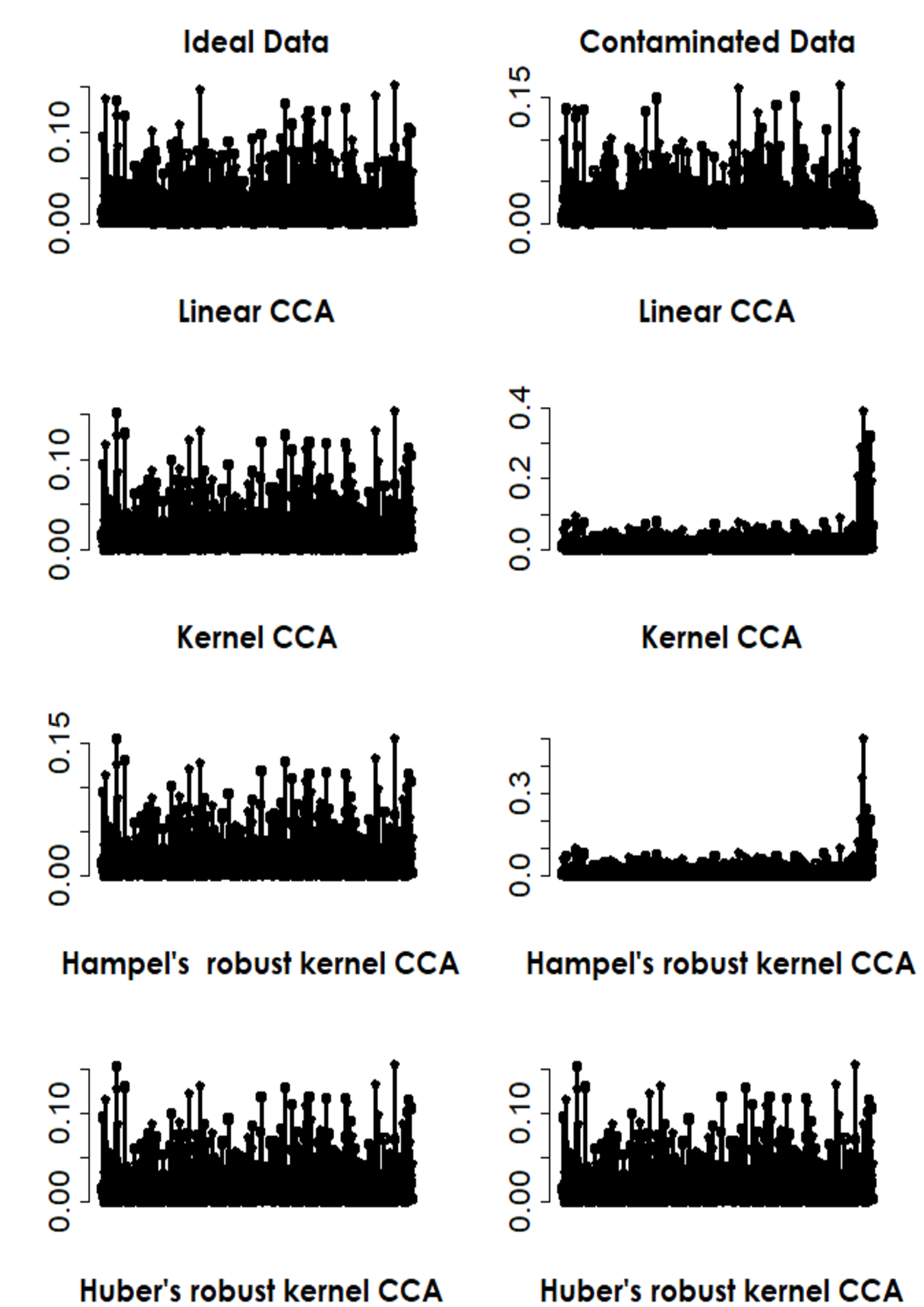}
  \caption{Comparison of influence value profiles across multiple kernel CCA methods for ideal and contaminated datasets using the RKUM package..}
  \label{figure:rlogo}
\end{figure}

\begin{figure}[htbp]
  \centering
  \includegraphics[width=\linewidth, height=0.9\textheight, keepaspectratio]{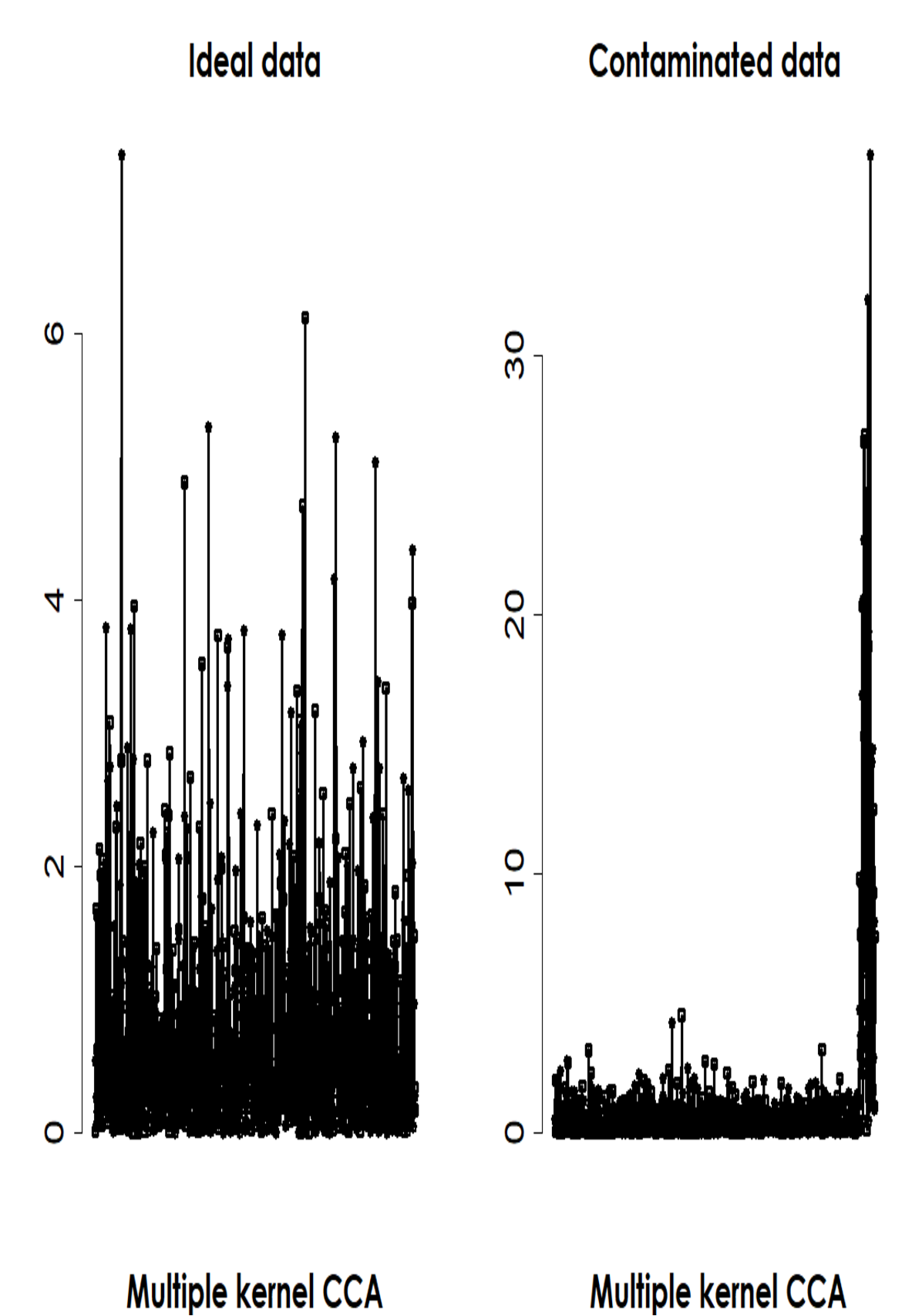}
  \caption{Line plots of influence values obtained from multiple kernel CCA methods applied to ideal and contaminated datasets using the RKUM package.}
  \label{figure:rlogo2}
\end{figure}

Figure 1 presents line graphs of the influence values estimated using kernel CCA methods for both ideal and contaminated data generated with the RKUM package. The results demonstrate that robust kernel methods employing robust loss functions—particularly Hampel’s and Huber’s loss functions—maintain stable influence profiles across components, indicating resilience to noise and contamination.

Figure 2 shows line plots of influence values obtained from multiple kernel CCA methods for both ideal and contaminated data. Under noisy conditions, the robust kernel CCA approaches implemented in the RKUM package exhibit reduced sensitivity to outliers compared with the standard kernel CCA, demonstrating improved robustness and stability across components.

\section{Summary}
\label{sec:conld}
The RKUM package provides an integrated computational framework for robust kernel-based unsupervised learning and canonical correlation analysis (CCA). It implements core statistical functions for estimating kernel covariance (CO) and kernel cross-covariance (CCO) matrices, along with influence function (IF) analysis for both standard and robust kernel CCA. The package supports multiple kernel types (e.g., linear, RBF, IBS) and incorporates robust loss functions such as Huber’s and Hampel’s, which are efficiently optimized using a Kernel Iteratively Reweighted Least Squares (KIRLS) approach.
Through these implementations, RKUM allows researchers to evaluate sensitivity, robustness, and influence of kernel-based estimators under both ideal and contaminated data conditions. It also provides visualization tools to identify influential or outlier data points in simulated and real-world multimodal datasets, including imaging-genetic data. Overall, RKUM offers a versatile and reliable platform for developing, testing, and interpreting robust kernel learning methods in high-dimensional data analysis.

\bibliographystyle{plainnat}

\bibliography{Ref-CI-2021}

\end{document}